\documentclass[10pt]{article}
\usepackage[utf8]{inputenc}
\usepackage[T1]{fontenc}
\usepackage{lmodern}
\usepackage[margin=0.92in]{geometry}
\usepackage{amsmath,amssymb}
\usepackage{booktabs,array,tabularx}
\usepackage{xcolor}
\usepackage{graphicx}
\usepackage{tikz}
\usetikzlibrary{arrows.meta,calc}
\usepackage{microtype}
\usepackage{natbib}
\usepackage{caption}
\usepackage{enumitem}
\usepackage{needspace}
\usepackage{placeins}
\definecolor{citeblue}{RGB}{70,140,200}
\usepackage[colorlinks=true,linkcolor=black,citecolor=citeblue,urlcolor=citeblue]{hyperref}
\setlist[itemize]{nosep,leftmargin=*}

\definecolor{depthblue}{RGB}{43,87,125}
\definecolor{timegreen}{RGB}{39,109,92}
\definecolor{panelgray}{RGB}{246,247,249}
\newcommand{\lrt}{\textnormal{\textsc{LRT}}}

\newcommand\blfootnote[1]{%
  \begingroup
  \renewcommand\thefootnote{}\footnote{#1}%
  \addtocounter{footnote}{-1}%
  \endgroup
}

\title{Trading Depth for Time in Recurrent Transformers}
\author{Zeyi Huang$^{1,2\star}$, Xuehai He$^{2\star}$, Yelong Shen$^{2\dagger}$,
Yong Jae Lee$^{1\dagger}$\\[0.6em]
$^1$University of Wisconsin--Madison \qquad $^2$Microsoft}
\date{}
\begin{document}
\maketitle

\begin{abstract}
Recurrent Transformers increase computational depth through temporal recurrence, feeding each token's high-level hidden state into the computation of the next. This raises a natural question: is additional computation better spent on more temporal steps or greater physical depth? We investigate this question using Latent Recurrent Transformers (LRT), which retain one backbone forward pass per vocabulary token during decoding and provide a controlled setting for comparing these two ways of adding computation. We insert a latent thought token between consecutive vocabulary tokens, adding a hidden-state refinement step through the shared backbone before predicting the next token.
On 16- and 20-layer NanoChat MoE backbones, we compare this model with double-depth vanilla Transformers and LRT, two-loop Transformers, and PonderLM-2 with one thought token. All configurations execute $2L$ blocks per vocabulary token during decoding. Under matched training FLOPs, LRT with one thought token outperforms PonderLM-2 and overtakes the two-loop Transformer as compute increases. Its gap to vanilla double-depth narrows from 0.014 to 0.001 bits per byte, despite approximately 46\% fewer parameters and approximately half as many unique training tokens. At the training budget used for our main comparison, one thought token recovers an average of 89\% of the gains from doubling LRT depth across the two backbone sizes, with approximately 48\% fewer parameters.
These results highlight temporal thinking as a parameter-efficient way to extend computation in recurrent Transformers.
\end{abstract}

\section{Introduction}
\blfootnote{$^\star$ and $^\dagger$ denote equal contribution and advising, respectively. Code is available at \url{https://github.com/OoDBag/Latent_recurrent_transformer}.}

\begin{figure}[t]
 \centering
 \IfFileExists{LRT_TT_2.png}{\includegraphics[width=\textwidth]{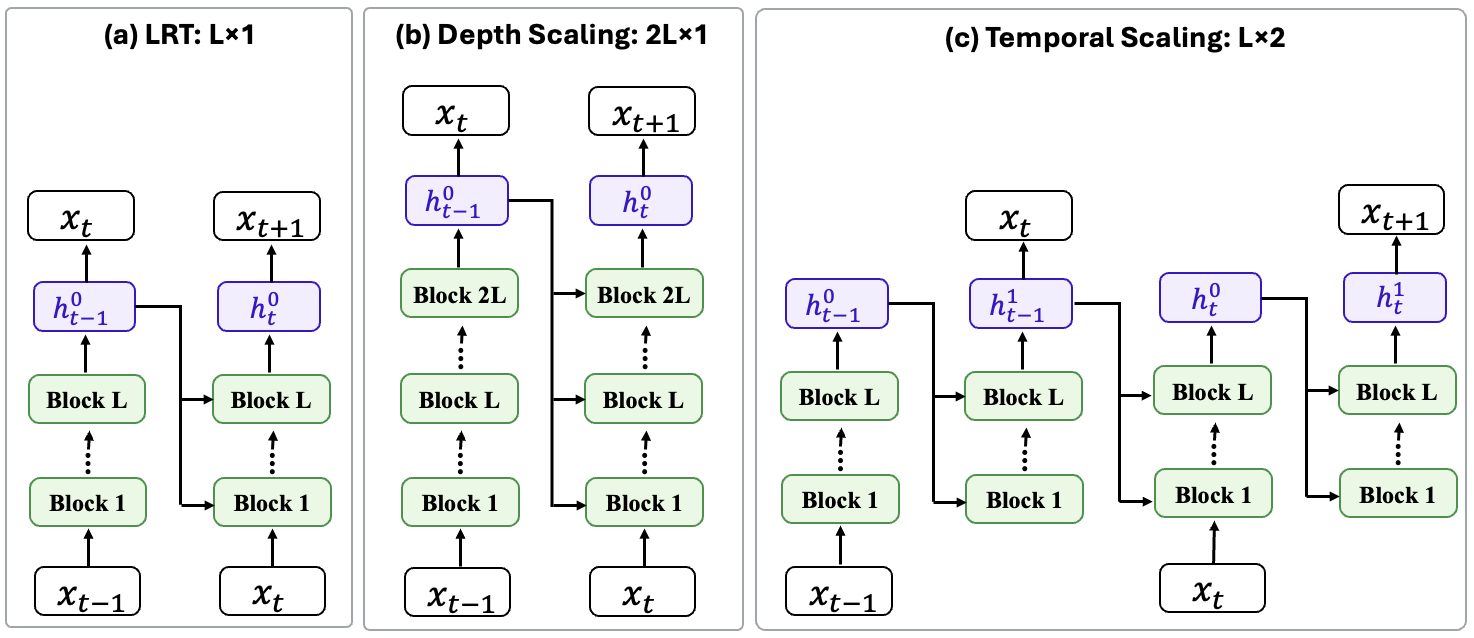}}{\fbox{\parbox[c][1.0in][c]{0.92\linewidth}{\centering Original architecture figure: \texttt{LRT\_TT\_2.png}\\Place the image beside this source to display it.}}}
 \caption{\textbf{Scaling LRT along depth or time.}
(a) An $L$-layer LRT performs one backbone forward pass per vocabulary token and feeds its final hidden state into the next token's computation.
(b) Depth scaling increases the backbone to $2L$ independently parameterized layers.
(c) Temporal scaling inserts one thought token after each vocabulary token, reusing the same $L$-layer backbone.
The vocabulary-token state $h_t^{(0)}$ is refined into the thought state $h_t^{(1)}$, which predicts $x_{t+1}$ and supplies feedback to its computation, continuing the recurrent chain across token boundaries.}
 \label{fig:architecture}
\end{figure}

The computational depth of a Transformer~\citep{vaswani2017attention} can be increased in several ways. Increasing physical depth adds independently parameterized layers, allowing each token to undergo more transformations before a prediction. Looped Transformers instead repeatedly apply a shared stack of layers to refine representations at the same token positions~\citep{dehghani2019universal,giannou2023looped,geiping2026scaling}. When unrolled, an $L$-layer backbone with $R$ loops executes $RL$ layers per token while retaining the same backbone parameters. Both approaches extend computation along the depth axis, differing primarily in whether successive transformations use distinct or shared parameters.

A different class of recurrent Transformers~\citep{fan2021feedback,liu2026maglev,cai2026t,huang2026latent} extends computation along the temporal axis by feeding a high-level hidden state from one token into the computation of the next. This cross-token feedback creates paths that repeatedly traverse the backbone as the sequence unfolds. The next token's computation can build directly on the preceding token's completed representation, so a dependency path can continue across vocabulary-token boundaries. This raises a complementary scaling question: rather than increasing the number of layers within each token's computation, can we obtain similar benefits by introducing additional steps along this recurrent sequence?

We study temporal computation using Latent Recurrent Transformers (LRT)~\citep{huang2026latent}, which reuse the preceding token's high-level hidden state through recurrent feedback. During autoregressive decoding, that state is already available, allowing LRT to retain one backbone forward pass per vocabulary token. LRT therefore provides a natural starting point for studying how to allocate additional computation: we can introduce more temporal steps or increase physical depth while keeping the recurrent architecture fixed.

Specifically, we insert a latent thought token between consecutive vocabulary tokens, providing another stage of hidden-state refinement before predicting the next token. The thought token passes through the same backbone and shares its parameters with the vocabulary-token computation. Its final state then supplies feedback to the next vocabulary token, preserving the cross-token recurrent chain.

On 16- and 20-layer NanoChat MoE backbones, we compare several ways of extending computation. Double-depth vanilla Transformers and LRT add independently parameterized layers, while looped Transformers extend depth by repeatedly applying shared layers. PonderLM-2~\citep{zeng2025ponderlm} and our method instead introduce additional temporal thinking steps. The configurations in our scaling comparison all execute $2L$ blocks per vocabulary token during decoding. Under matched training FLOPs, LRT with one thought token outperforms PonderLM-2 and overtakes a two-loop Transformer as training compute increases. Its gap to a vanilla double-depth Transformer narrows from 0.014 to 0.001 BPB, despite using approximately 46\% fewer parameters and half as many unique training tokens. In the main comparison (Table~\ref{tab:main}), one thought token recovers an average of 89\% of the gains from doubling LRT depth across the two backbone sizes, with approximately 48\% fewer parameters.

To understand why our temporal extension outperforms PonderLM-2, we examine how computation continues across vocabulary-token boundaries. Both methods introduce thought steps before prediction, but they differ in cross-token feedback: LRT passes the final thought state directly into the next vocabulary token's computation, whereas PonderLM-2 does not. This allows LRT's recurrent dependency chain to continue across tokens. Removing this connection reduces performance toward PonderLM-2. This ablation supports cross-token recurrence as a key contributor to the gains, beyond the additional computation provided by thought steps alone.

Finally, sampling the number of thought tokens during training allows one model to use different numbers of thought steps at inference and improves performance over training with fixed thought steps within the trained range. This also lays the groundwork for future adaptive computation that allocates different numbers of thought steps to individual tokens.

Together, these experiments examine both the depth--time trade-off and the feedback connections that support temporal computation.

\section{Method}
\label{sec:method}

Starting from an $L$-layer Latent Recurrent Transformer (LRT), we introduce additional temporal computation through latent thought tokens that reuse the backbone. We first describe the recurrent backbone and thought-token construction, then explain computation accounting and training with variable thinking steps. Figure~\ref{fig:architecture} illustrates temporal scaling alongside physical-depth scaling.

\subsection{Latent Recurrent Backbone}
\label{sec:lrt_backbone}

Let $x_1,\ldots,x_N$ be a sequence of vocabulary tokens, and let $E[x_t]$ denote the embedding of token $x_t$. LRT passes the preceding token's top-layer hidden state into the current token's computation:
\begin{equation}
 h_t=F_\theta\bigl(E[x_t],h_{t-1};C_{<t}\bigr),
 \label{eq:lrt_backbone}
\end{equation}
where $F_\theta$ denotes the backbone together with its feedback pathways, $h_{t-1}$ is the preceding token's final hidden state, and $C_{<t}$ contains the earlier positions' KV states. Each forward pass also adds the current position's KV states to the cache; we omit those updates from the equation for readability. The model predicts $x_{t+1}$ from $h_t$ using its output head.

This feedback allows the current prediction to build on a representation that has already passed through the backbone at the preceding token. Since $h_{t-1}$ is available during autoregressive decoding, the recurrence requires only one new backbone forward pass per vocabulary token.

\paragraph{Feedback configuration.}

We use the final-layer hidden state as the recurrent source and inject it through hidden-state residuals and KV feedback, following \citet{huang2026latent}. Learned scales and gates control these contributions. We keep this feedback design fixed across the model variants studied below. The feedback projections add approximately 5\% to the parameter count of the corresponding vanilla backbone.

\subsection{Adding Thought Tokens}
\label{sec:temporal_thinking}

We insert continuous thought positions after each vocabulary token. Each thought position takes the preceding hidden state as its input and processes it through the shared backbone, without sampling a discrete vocabulary token. The final thought state predicts the next vocabulary token and supplies recurrent feedback to its computation.

More generally, let $h_t^{(0)}$ be the state after processing $x_t$, and $h_t^{(j)}$ the state after processing its $j$th thought token. With $K$ thought tokens per vocabulary token,
\begin{align}
 h_t^{(0)}&=F_\theta\bigl(E[x_t],h_{t-1}^{(K)};C_{<t,0}\bigr),
 \label{eq:tokenstep}\\
 h_t^{(j)}&=F_\theta\bigl(h_t^{(j-1)},h_t^{(j-1)};C_{<t,j}\bigr),
 \label{eq:thinkstep}\\
 &\hspace{4em}j=1,\ldots,K.\notag
\end{align}
Here $C_{<t,j}$ contains all positions preceding the current step. The two occurrences of $h_t^{(j-1)}$ in Eq.~\eqref{eq:thinkstep} have different roles: the state replaces a token embedding at the input and also supplies the recurrent feedback within the backbone. The original output head reads the final state $h_t^{(K)}$ to predict $x_{t+1}$. Setting $K=0$ recovers LRT without thought tokens.

\paragraph{Parameter sharing and attention.}
Thought tokens have no learned token embedding. Vocabulary and thought positions share the attention, MoE, and feedback-projection parameters; only their feedback scales and gates are separate. The same feedback projections are reused across different thought steps, so increasing the number of thought steps does not replicate their parameter overhead. Adding thought tokens requires only a small number of additional scales and gates, making its incremental parameter cost negligible.

The MoE router can select different experts across vocabulary and thought positions, allowing successive steps to activate different subsets of the shared parameters.

All positions write KV states and remain visible to later positions under the attention mask. Thought tokens share their associated vocabulary token's rotary position index: for $K=1$, the indices are $0,0,1,1,\ldots$, and for $K=2$, they are $0,0,0,1,1,1,\ldots$.

\paragraph{Cross-token feedback.}
The key distinction from PonderLM-2 is how computation continues after a thought token. LRT with thought tokens feeds the final thought state directly into the next vocabulary token's computation, allowing the recurrent dependency chain to continue across vocabulary-token boundaries. Without this connection, the explicit feedback chain ends at the thought token, although information remains accessible through causal attention. Section~\ref{sec:ablation} examines the contribution of this cross-token connection.

\subsection{Comparing Physical Depth and Temporal Computation}
\label{sec:accounting}

We hold width and the recurrent feedback design fixed when comparing physical-depth and temporal scaling within LRT. An $L$-layer model with $K$ thought tokens executes
\begin{equation}
 B_{\mathrm{decode}}(L,K)=L(K+1)
 \label{eq:decodework}
\end{equation}
blocks per vocabulary token during decoding. Thus, one thought token matches the decoding block count of doubling depth, and two match that of tripling depth. The deeper model applies additional distinct layers, whereas the temporal model repeatedly applies the shared backbone. For looped models, the decoding block count is the physical layer count multiplied by the number of loops.

This count measures newly executed blocks per vocabulary token, not the depth of the full computational graph. Computational depth counts block transformations along the longest sequential dependency path to a prediction. Cross-token feedback allows this path to continue through earlier tokens, so equal decoding block counts need not imply equal computational depth.
\paragraph{Parameters and attention cost.}
Thought tokens add only a negligible number of feedback scales and gates, while depth scaling adds full backbone layers. Doubling physical depth need not exactly double total parameters, since token embeddings and the output projection do not scale with the number of layers.

Equal block counts also do not imply equal FLOPs or latency. With full causal attention and equal KV dimensions, a $2L$-layer model over $N$ vocabulary positions and an $L$-layer model over $2N$ vocabulary-and-thought positions both store $2LN$ layer--position KV entries. However, the thought-token model attends over more positions. Training-compute comparisons must also account for refinement passes, backward computation, and the expanded attention sequence. We report decoding latency in Section~\ref{sec:experiments} and summarize where each method spends training compute in Table~\ref{tab:compute_breakdown} (Appendix~\ref{app:compute}).

\begin{figure}[t]
 \centering
 \includegraphics[width=0.8\linewidth]{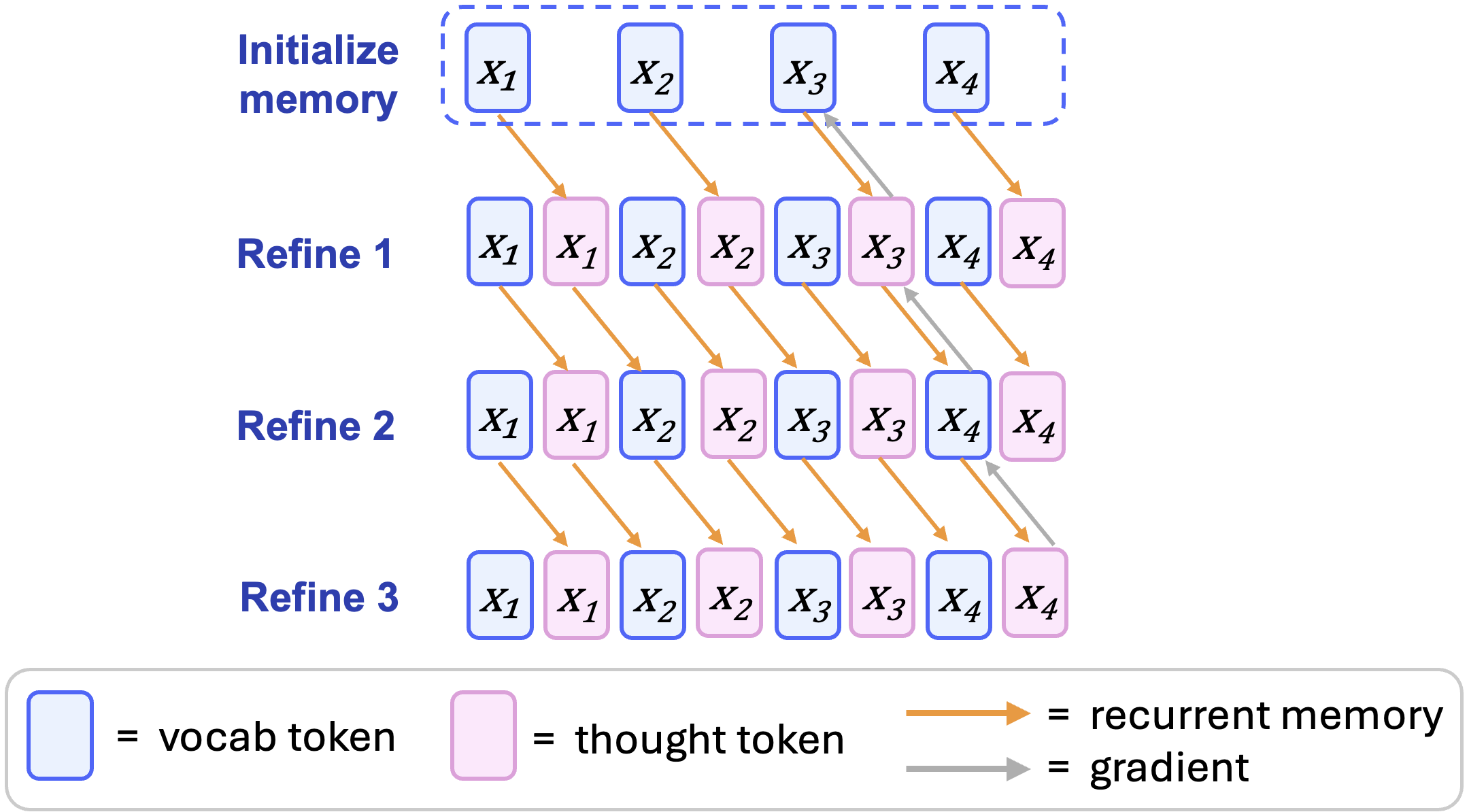}
 \caption{
 \textbf{Parallel multi-refinement training and gradient flow.} A vocabulary-only initialization pass is followed by three refinement passes in this example ($P=3$).
 Each box represents one forward pass through the shared $L$-layer backbone.
 Orange arrows pass final hidden states from the preceding pass to the next position in the expanded sequence.
 Gray arrows highlight the recurrent gradient path from the thought state associated with $x_4$ in Refine~3, through vocabulary token $x_4$ in Refine~2, the thought state associated with $x_3$ in Refine~1, and vocabulary token $x_3$ in initialization.
 This illustrated path traverses four backbone evaluations across two vocabulary-token intervals.
 Attention connections and other gradient paths are omitted.
 }
 \label{fig:training}
\end{figure}

\subsection{Parallel Multi-Refinement Training}
\label{sec:training}

The equations above describe sequential decoding, where each completed state is available to the next position. Applying the same recurrence exactly during training would require each position to wait for the preceding position's final state, preventing a straightforward parallel forward over the sequence. We instead use multi-refinement training~\citep{zeng2025ponderlm,huang2026latent}: each pass processes all positions in parallel, taking recurrent inputs from the preceding pass's final states shifted by one position. For thought-token models, these positions belong to the expanded vocabulary-and-thought sequence. At thought positions, the shifted state also supplies the input embedding replacement. Figure~\ref{fig:training} illustrates this procedure.

\paragraph{Training procedure.}

LRT thought-token models begin with a vocabulary-only initialization pass using zero recurrent input and no direct prediction loss, followed by refinement passes over $N(K+1)$ positions. LRT without thought tokens refines the vocabulary sequence. Across all LRT experiments, we sample one refinement count per batch, $P\in\{1,2,3,4\}$ with probabilities $(0.4,0.3,0.2,0.1)$, respectively, giving $\mathbb{E}[P]=2.0$. For thought-token models, $P$ excludes the initialization pass.

We average next-token loss equally over the supervised passes and retain gradients across passes. Each vocabulary-token interval predicts from its final state: the last thought state when $K>0$, or the vocabulary state when $K=0$. PonderLM-2 uses the same refinement-count distribution, also giving two refinement passes in expectation, while looped Transformers unroll repeated backbone applications directly in the training graph.

\begin{figure}[t]
    \centering
    \includegraphics[width=0.8\linewidth]{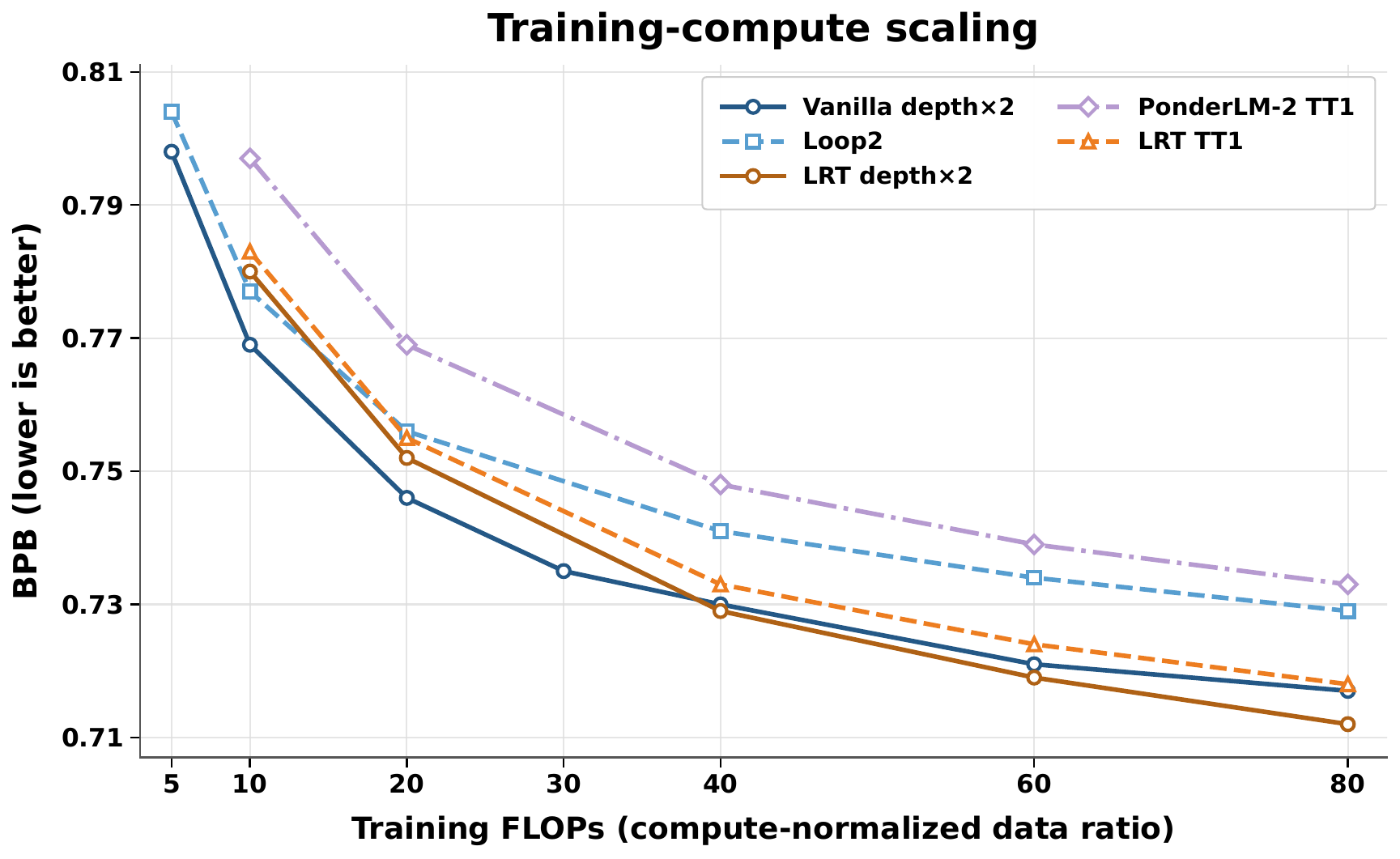}
    \caption{
    \textbf{Performance versus training compute.}
    Validation BPB ($\downarrow$) versus normalized training FLOPs.
    The compute-normalized data ratio expresses unique training tokens per reference parameter, adjusted for each method's compute cost (Appendix~\ref{app:results}).
    \textbf{At matched training FLOPs, the vanilla double-depth
    Transformer and Loop2 use approximately twice as many unique
    training tokens as double-depth LRT, PonderLM-2 with one thought
    token, and LRT with one thought token.}
    LRT with one thought token outperforms PonderLM-2 and overtakes
    Loop2 as compute increases, while its gap to the vanilla
    double-depth Transformer narrows to 0.001 BPB.
    Lines connect observed results without fitting a scaling law.
    }
    \label{fig:compute_scaling}
\end{figure}

\paragraph{Variable thinking steps.}
We sample one $K$ per batch, uniformly from $\{0,1\}$ for LRT TT1 and from
$\{0,1,2\}$ for LRT TT2, rather than fixing $K=1$ or $K=2$
throughout training. The expected thought counts are therefore
$0.5$ and $1$, respectively. In fixed--variable comparisons, we hold the expanded sequence length and optimizer-step count fixed within each pair. Smaller sampled $K$ values allow more vocabulary tokens per sequence; we revisit the same unique-data corpus to supply these additional exposures. Appendix~\ref{app:compute} gives the FLOP calculation and matching protocol.
At evaluation, the same weights support different numbers of thought steps;
unless otherwise stated, LRT TT1 and TT2 use $K=1$ and $K=2$.
This flexibility provides a foundation for future adaptive
allocation of thought steps across tokens.

\section{Experiments}
\label{sec:experiments}

\begin{table}[!t]
\centering
\small
\setlength{\tabcolsep}{3pt}
\renewcommand{\arraystretch}{1.12}
\caption{
\textbf{Quality and parameter trade-offs at data ratio 10.}
Models use the same training steps and unique vocabulary-token
data budget within each reference size (Appendix~\ref{app:compute}). BPB is averaged over three
seeds. Physical depth and total parameters are normalized to the corresponding vanilla reference model. The single decoding-latency column reports the arithmetic mean of the normalized latencies at the 16L and 20L reference sizes, each normalized to its own vanilla baseline.
Training FLOPs are not matched across all rows;
selected pairs, including double-depth LRT and LRT with one
thought token, have approximately matched training FLOPs.
The one- and two-thought-token LRT are trained with
$K\in\{0,1\}$ and $K\in\{0,1,2\}$, respectively, and evaluated
at their largest trained numbers of thought steps.
Bold and underlined values indicate the best and second-best BPB
within the approximately $1\times$, $2\times$, and $3\times$
latency groups, separately for each model size.
}
\label{tab:main}
\begin{tabular*}{\linewidth}{@{\extracolsep{\fill}}lcccccc@{}}
\toprule
 & & & \multicolumn{2}{c}{Total params}
 & \multicolumn{2}{c}{BPB $\downarrow$}\\
\cmidrule(lr){4-5}\cmidrule(l){6-7}
Model & Physical depth & \shortstack{Decode\\latency}
& 16L & 20L & 16L & 20L\\
\midrule
Transformer
& $1\times$ & $1.0\times$
& $1.00\times$ & $1.00\times$
& \underline{0.818} & \underline{0.788}\\
Transformer
& $2\times$ & $2.0\times$
& $1.92\times$ & $1.94\times$
& 0.798 & 0.767\\
Transformer
& $3\times$ & $3.0\times$
& $2.83\times$ & $2.87\times$
& 0.789 & 0.756\\
\midrule
Loop Transformer, 2 loops
& $1\times$ & $2.0\times$
& $1.00\times$ & $1.00\times$
& 0.804 & 0.774\\
Loop Transformer, 3 loops
& $1\times$ & $3.0\times$
& $1.00\times$ & $1.00\times$
& 0.799 & 0.768\\
Loop Transformer, 4 loops
& $1\times$ & $3.9\times$
& $1.00\times$ & $1.00\times$
& 0.795 & 0.764\\
\midrule
PonderLM-2, 1 thought token
& $1\times$ & $2.1\times$
& $1.00\times$ & $1.00\times$
& 0.797 & 0.752\\
PonderLM-2, 2 thought tokens
& $1\times$ & $3.1\times$
& $1.00\times$ & $1.00\times$
& 0.789 & 0.745\\
\midrule
\lrt
& $1\times$ & $1.1\times$
& $1.04\times$ & $1.05\times$
& \textbf{0.798} & \textbf{0.762}\\
\lrt
& $2\times$ & $2.2\times$
& $2.00\times$ & $2.04\times$
& \textbf{0.780} & \textbf{0.741}\\
\lrt
& $3\times$ & $3.3\times$
& $2.96\times$ & $3.03\times$
& \textbf{0.771} & \textbf{0.731}\\
\midrule
\lrt + 1 thought token
& $1\times$ & $2.3\times$
& $1.04\times$ & $1.05\times$
& \underline{0.783} & \underline{0.742}\\
\lrt + 2 thought tokens
& $1\times$ & $3.4\times$
& $1.04\times$ & $1.05\times$
& \underline{0.776} & \underline{0.736}\\
\bottomrule
\end{tabular*}
\end{table}

\subsection{Setup}
\label{sec:setup}

\paragraph{Models.}
We evaluate NanoChat~\citep{karpathy2025nanochat} MoE backbones with reference depths of
16 and 20 layers, following \citet{huang2026latent}. Unless stated otherwise, we inherit the data, optimization, and evaluation settings of that work.
Depth multipliers of $2\times$ and $3\times$ correspond to
32/48 and 40/60 layers, respectively, with width fixed
within each reference size.
The MoE configuration uses eight routed experts, top-2 routing,
and one shared expert.
The reference vanilla Transformers contain 0.8B and 1.3B
total parameters. Thought-token and looped models retain
the reference backbone depth.

\paragraph{Training.}
All LRT experiments sample one refinement count per batch, $P\in\{1,2,3,4\}$ with probabilities $(0.4,0.3,0.2,0.1)$ and $\mathbb{E}[P]=2.0$.
We average next-token loss over supervised refinement passes
and retain gradients across passes.
LRT TT1 and TT2 uniformly sample $K\in\{0,1\}$
and $K\in\{0,1,2\}$, respectively. PonderLM-2 uses the same $P$ distribution. Fixed--variable comparisons match training FLOPs separately within each pair by using the same expanded sequence length, batch size in sequences, and number of updates; both runs reuse the same unique-data corpus (Appendix~\ref{app:compute}).

\paragraph{Evaluation.}
We report validation bits per byte (BPB; lower is better).
Evaluation executes sequential recurrence with KV caching:
the model receives each ground-truth vocabulary token and
scores the next one from the final state of its interval.
Unless otherwise stated, LRT TT1 and TT2 are evaluated with
$K=1$ and $K=2$, respectively.
Table~\ref{tab:variable_thinking} additionally evaluates each
model at other numbers of thought steps using the same weights.
The training refinement count $P$ is not an additional
iteration count at decoding time.

\paragraph{Budgets and reporting.}
We use the vanilla double-depth Transformer as the reference for data-ratio normalization and training-compute budgets. Table~\ref{tab:main} reports results at data ratio 10,
holding training steps and the unique vocabulary-token data budget
fixed within each reference size. It compares quality and
parameters at similar decoding latencies without imposing
a common training-FLOP budget across all rows.
For each scaling budget in Figure~\ref{fig:compute_scaling}, we divide the reference training-FLOP budget by each model's expected FLOPs per optimizer step to determine its step count. The calculation uses the model's depth, refinement schedule, and expanded attention sequence, rather than a sequence-length multiplier alone. Appendix~\ref{app:compute} gives the formulas and explains how this comparison differs from the pairwise fixed--variable protocol.
Appendix~\ref{app:details} provides the data-ratio definition,
training settings, compute accounting, and latency measurement
details; Appendix~\ref{app:results} reports the numerical
scaling results. Decoding latency follows the measurement protocol of \citet{huang2026latent}. Appendix~\ref{app:std} describes seed-variability reporting for Table~\ref{tab:main}.

\subsection{Scaling with Training Compute}
\label{sec:compute_scaling}

Figure~\ref{fig:compute_scaling} compares double-depth vanilla
Transformers and LRT, two-loop Transformers (Loop2), PonderLM-2 with one thought token,
and LRT with one thought token.
All execute $2L$ Transformer blocks per vocabulary token
during decoding.
\textbf{To match training FLOPs, the vanilla double-depth
Transformer and Loop2 use approximately twice as many unique
training tokens as the other three models.}
See Appendix~\ref{app:compute} for the budget calculation and Appendix~\ref{app:results} for the numerical values.

LRT with one thought token consistently outperforms PonderLM-2
and overtakes Loop2 as training compute increases.
At the largest reported matched budget, it reaches 0.718 BPB,
compared with 0.733 for PonderLM-2 and 0.729 for Loop2.
The vanilla double-depth Transformer remains stronger at
0.717 BPB, but its advantage narrows from 0.014 to 0.001 BPB
across the tested budgets.

The comparison with double-depth LRT shows a different trend.
Its advantage over LRT with one thought token increases from
0.003 to 0.006 BPB between the smallest and largest reported
data budgets.
Temporal thinking therefore offers a strong parameter--quality
trade-off, while additional independently parameterized layers
retain an advantage as training data increases.

\subsection{Trading Parameters for Temporal Computation}

At data ratio 10 (Table~\ref{tab:main}), one thought token
reaches 0.783 and 0.742 BPB at the two reference sizes,
compared with 0.780 and 0.741 for double-depth LRT.
It recovers 83.3\% and 95.2\% of the improvement from doubling
LRT depth with approximately 48\% fewer total parameters and
similar decoding latency ($2.3\times$ versus $2.2\times$).
These recovery rates describe this training budget rather
than a constant advantage across data scales.
Appendix~\ref{app:recovery} defines the recovery metric.

\subsection{Further Gains from Additional Thought Steps}

Increasing the maximum number of thought steps during training from one to two improves BPB from 0.783 to 0.776 at the 16-layer reference
size and from 0.742 to 0.736 at the 20-layer size, when each
model is evaluated using its largest trained number of thought steps.
Both models share the same backbone size, while the two-thought
model executes three backbone forward passes per vocabulary token.

At the same decoding block count, triple-depth LRT models reach
0.771 and 0.731 BPB.
Two thought tokens therefore recover 81.5\% and 83.9\% of the
improvement from tripling LRT depth, leaving a 0.005 BPB gap
at both sizes while using approximately 65\% fewer parameters.
Their relative decoding latency is $3.4\times$, compared with
$3.3\times$ for triple-depth LRT.
Thus, the parameter--quality trade-off extends beyond a single
additional thought step.

These comparisons use separately trained models.
Table~\ref{tab:variable_thinking} also isolates additional
inference computation within one model: for weights trained
with $K\in\{0,1,2\}$, increasing the number of thought steps at evaluation from
$K=0$ to $K=1$ and $K=2$ improves BPB from 0.789 to
0.778 and 0.776, respectively.

\begin{table}[!t]
\centering
\small
\setlength{\tabcolsep}{2.5pt}
\renewcommand{\arraystretch}{1.12}
\caption{
\textbf{Fixed versus variable thinking steps at depth 16.}
Validation BPB ($\downarrow$) at evaluation thought count $K$.
Each fixed--variable pair uses the same expanded sequence length, update count, and unique-data corpus to match expected training FLOPs. Vocabulary-token exposure may differ because the corpus is revisited. Budgets are matched pairwise, not globally across all rows (Appendix~\ref{app:compute}).
Each row represents one trained model; rows trained with sampled $K$ reuse
the same weights across columns.
$\dagger$ marks evaluation beyond that row's training range.
Dashes indicate unreported evaluations.
}
\label{tab:variable_thinking}
\begin{tabularx}{0.80\linewidth}{@{}>{\raggedright\arraybackslash}Xccccc@{}}
\toprule
& \multicolumn{5}{c}{Evaluation thought steps}\\
\cmidrule(lr){2-6}
Training
& $K=0$ & $K=1$ & $K=2$ & $K=3$ & $K=4$\\
\midrule
Fixed $K=0$
& 0.798 & -- & -- & -- & --\\
Fixed $K=1$
& -- & 0.786 & -- & -- & --\\
Fixed $K=2$
& -- & -- & 0.779 & -- & --\\
Fixed $K=3$
& -- & -- & -- & 0.779 & --\\
Fixed $K=4$
& -- & -- & -- & -- & 0.779\\
\midrule
One model, sampled $K\in\{0,1\}$
& 0.794
& 0.783
& 0.792$^{\dagger}$
& 0.809$^{\dagger}$
& 0.835$^{\dagger}$\\
One model, sampled $K\in\{0,1,2\}$
& 0.789
& 0.778
& 0.776
& 0.782$^{\dagger}$
& 0.787$^{\dagger}$\\
\bottomrule
\end{tabularx}
\end{table}

\subsection{Training with Variable Thinking Steps}
\label{sec:variable_thinking}

Table~\ref{tab:variable_thinking} compares training with fixed and variable thought steps at depth 16. Each model trained with sampled $K$ is matched to its counterpart trained at fixed $K=K_{\max}$ using the same expanded sequence length, batch size in sequences, refinement-count distribution, and optimizer-step count. Both runs use the same unique-data corpus, revisiting it as needed. Appendix~\ref{app:compute} details this pairwise matching.
For LRT TT1, sampling $K\in\{0,1\}$ improves BPB at $K=1$
from 0.786 to 0.783 compared with fixed-$K=1$ training.
For LRT TT2, sampling $K\in\{0,1,2\}$ improves BPB at $K=2$
from 0.779 to 0.776 compared with fixed-$K=2$ training.
Both models trained with sampled $K$ also have lower BPB than the separately trained references trained with fixed $K$ at their lower trained values of $K$. These results at other values of $K$ demonstrate inference flexibility; they are not additional pairwise FLOP-matched training comparisons. Within-pair gains therefore compare fixed and sampled numbers of thought steps at the same update count and unique-data budget, although their cumulative vocabulary-token exposures can differ.

Additional evaluation steps do not improve BPB beyond the trained range:
the model trained with $K\in\{0,1,2\}$ degrades from
0.776 BPB at $K=2$ to 0.782 and 0.787 at $K=3,4$.
Generalization to more thought steps and adaptive
per-token allocation remain open.

\begin{table}[!t]
\centering
\small
\setlength{\tabcolsep}{3pt}
\renewcommand{\arraystretch}{1.12}
\caption{
\textbf{Feedback and training ablations at depth 16 with one
thought token.}
All ablations use data ratio 10. Each independently removes one component from the
full LRT configuration.
PonderLM-2 is included as a reference without explicit
cross-token recurrent feedback.
}
\label{tab:ablation}
\begin{tabularx}{0.70\linewidth}{@{}>{\raggedright\arraybackslash}Xcc@{}}
\toprule
Configuration & BPB $\downarrow$
& \shortstack{Increase over\\full LRT}\\
\midrule
\lrt + 1 thought token
& 0.783 & ---\\
\quad Without cross-token recurrence
& 0.792 & +0.009\\
\quad Without KV feedback
& 0.787 & +0.004\\
\quad Without thinking step sampling
& 0.786 & +0.003\\
\midrule
PonderLM-2, 1 thought token
& 0.797 & +0.014\\
\bottomrule
\end{tabularx}
\end{table}

\subsection{Ablating Recurrent Feedback and Variable Thinking Steps}
\label{sec:ablation}

Table~\ref{tab:ablation} independently ablates cross-token
recurrence, KV feedback, and training with variable thought steps in the
16-layer LRT at data ratio 10, evaluated with one thought token.

\paragraph{Cross-token recurrence and PonderLM-2.}
Both LRT with thought tokens and PonderLM-2 introduce temporal
thinking steps.
LRT additionally carries the final thought state into the next
vocabulary token through explicit recurrent feedback.
Removing this connection preserves thought steps and causal
attention but increases BPB from 0.783 to 0.792, substantially
narrowing the gap to PonderLM-2's 0.797.
This supports cross-token recurrence as a contributor to the
gains beyond adding thought steps alone.

\paragraph{KV feedback and training with variable thought steps.}
Removing KV feedback increases BPB to 0.787.
Replacing training with variable thought steps with fixed-$K=1$ training
increases BPB to 0.786.
Both components therefore contribute to the full model's
performance. The cross-token ablation removes 0.009 BPB of the full model's 0.014 BPB advantage over PonderLM-2, leaving a residual 0.005 BPB advantage. It supports the contribution of the connection without identifying computational depth as the sole explanation for the gains.

\FloatBarrier

\section{Related Work}
\label{sec:related_work}

\paragraph{Depth recurrence and parameter sharing.}
Universal Transformers~\citep{dehghani2019universal} and looped Transformers~\citep{giannou2023looped} reuse transformations to increase computation without proportionally increasing parameters. Recurrent-depth language models extend this principle to latent test-time computation~\citep{geiping2026scaling}. We compare shared-depth computation with additional physical layers and temporal thought positions, accounting separately for training FLOPs and decoding cost.

\paragraph{Temporal feedback and recurrent memory.}
Feedback Transformers expose high-level past representations to subsequent computation~\citep{fan2021feedback}. T$^2$MLR introduces middle-layer temporal recurrence~\citep{cai2026t}, while Maglev combines recurrent KV memory with sliding-window decoding~\citep{liu2026maglev}. We build on LRT's hidden-state and KV feedback~\citep{huang2026latent}, keeping these pathways fixed to study how additional thought steps trade off against physical depth.

\paragraph{Continuous thought steps.}
Pause tokens introduce additional computation before output generation by inserting learned token embeddings~\citep{goyal2024think,herel2024thinking,pfau2024let}. PonderLM-2 inserts continuous thought states before vocabulary prediction and trains with parallel refinement~\citep{zeng2025ponderlm}. However, it does not directly feed the final thought state into the next vocabulary token's computation through recurrent feedback. LRT preserves this connection, allowing the refined state to support both the current prediction and subsequent token computation. This cross-token feedback distinguishes our temporal extension from PonderLM-2.

\section{Conclusion}

Temporal computation offers a parameter-efficient way to recover much of the benefit of additional physical depth in recurrent Transformers. Our results suggest that the value of extra computation depends on how its states are carried forward: thought steps refine the state used for prediction, while cross-token feedback passes that state into subsequent token computation. Ablations support this connection as a contributor to the gains beyond adding thought steps alone. Physical depth retains an advantage in our comparisons, so temporal steps provide a useful trade-off rather than a complete substitute. Training with variable thought steps further makes temporal computation adjustable within a single model over the trained range.

\section*{Limitations}

Our study examines the trade-off between physical depth and temporal computation on two NanoChat MoE backbone sizes, using language-modeling BPB as the primary metric. Whether the same trade-off holds for larger models, other architectures, and downstream tasks such as reasoning remains to be established. Although we evaluate multiple training budgets, the observed trends characterize the tested range and do not establish a general scaling law.

At inference, we assign the same number of thought steps to every vocabulary token. Training with variable thought steps allows one model to operate at several fixed numbers of thought steps, but does not learn to allocate computation according to the needs of individual tokens. Adaptive Computation Time~\citep{graves2016adaptive} learns input-dependent computation lengths. PonderLM-3~\citep{li2026ponderlm} learns token-wise adaptive pondering through differentiable attention masking and inference-time stopping. Incorporating such allocation and stopping mechanisms into our LRT framework, and evaluating their effects on prediction quality and realized latency, remains future work. Moreover, additional thought steps beyond the trained range do not consistently improve performance in our experiments, leaving reliable generalization to more thought steps at inference an open question.

\section*{Acknowledgments}

This work was supported in part by NSF IIS2404180, National Research Foundation of Korea (NRF) grant (MSIT) (RS-2026-25522067) and Institute of Information \& communications Technology Planning\& Evaluation (IITP) grants funded by the Korea government (MSIT) (No. 2022-0-00871, Development of AI Autonomy and Knowledge Enhancement for AI Agent Collaboration), (No. RS-2025-2543949. Environment-Aware and Domain-Adaptive Multimodal Embodied AI for Real-World Interaction), and Electronics and Telecommunications Research Institute (ETRI) grant (26CB1200, Development and Application of Science-Specialized Multimodal Foundation Models).

\bibliographystyle{plainnat}
\bibliography{main}

\appendix

\section{Additional Scaling Results}
\label{app:results}

Table~\ref{tab:scaling_full} reports the numerical results used in Figure~\ref{fig:compute_scaling}. These scaling runs read new corpus positions as training proceeds, so additional optimizer steps increase the number of unique training tokens. Following the total-parameter convention of \citet{huang2026latent}, we normalize the number of unique vocabulary-token occurrences $U_m$ by the total trainable parameter count $N_{\mathrm{ref}}$ of the vanilla double-depth Transformer for the corresponding reference size. Thus the table's data ratio is $D=U_m/N_{\mathrm{ref}}$, with one common denominator across methods. Inserted thought positions and repeated refinement passes over the same batch do not count as additional unique training tokens. Here ``unique'' refers to corpus token occurrences, not distinct vocabulary types.

The figure uses an approximate compute normalization: a unique-training-token ratio $D$ is plotted at $D/2$ for the vanilla double-depth Transformer and Loop2, and at $D$ for the other three models. Under this normalization, comparisons pair unique-training-token ratio $D$ for the recurrent models with $2D$ for the vanilla double-depth Transformer and Loop2. Training-step budgets are determined from average step FLOPs as described in Appendix~\ref{app:compute}; the factor-of-two display convention is not an exact FLOP identity for arbitrary architectures or sampling schedules.

\begin{table}[t]
\centering
\small
\setlength{\tabcolsep}{6pt}
\renewcommand{\arraystretch}{1.12}
\begin{tabular}{lccccccc}
\toprule
& \multicolumn{7}{c}{Unique-training-token ratio $D$}\\
\cmidrule(lr){2-8}
Model & 10 & 20 & 40 & 60 & 80 & 120 & 160\\
\midrule
Vanilla Transformer, depth $2\times$
& 0.798 & 0.769 & 0.746 & 0.735 & 0.730 & 0.721 & 0.717\\
LRT, depth $2\times$
& 0.780 & 0.752 & 0.729 & 0.719 & 0.712 & -- & --\\
Loop Transformer, 2 loops
& 0.804 & 0.777 & 0.756 & -- & 0.741 & 0.734 & 0.729\\
PonderLM-2, 1 thought token
& 0.797 & 0.769 & 0.748 & 0.739 & 0.733 & -- & --\\
LRT, 1 thought token
& 0.783 & 0.755 & 0.733 & 0.724 & 0.718 & -- & --\\
\bottomrule
\end{tabular}
\caption{
\textbf{Numerical scaling results.}
Validation BPB ($\downarrow$) across unique-training-token ratios $D$. Entries in the same column use the same number of unique training tokens, not necessarily the same training FLOPs. At matched training FLOPs, the vanilla double-depth Transformer and Loop2 use approximately twice as many unique training tokens as the other three models.
Dashes indicate unreported results.
}
\label{tab:scaling_full}
\end{table}

At the smallest reported matched compute budget, LRT with one
thought token at $D=10$ reaches 0.783 BPB, compared with 0.769
for the vanilla double-depth Transformer and 0.777 for Loop2
at $D=20$. At the largest matched budget, the corresponding
values are 0.718 at $D=80$, versus 0.717 and 0.729 at $D=160$.
Thus, the gap to the vanilla double-depth Transformer narrows
from 0.014 to 0.001 BPB, while the comparison with Loop2 changes
from a 0.006 BPB deficit to a 0.011 BPB advantage.

At equal unique-training-token budgets, double-depth LRT remains stronger than
LRT with one thought token. Their BPB gaps at
$D=10,20,40,60,80$ are 0.003, 0.003, 0.004, 0.005, and 0.006,
respectively. These results distinguish approaching a
vanilla model trained on more unique tokens at matched training compute from
recovering the gains of additional depth within LRT.

\section{Measuring Recovery of Physical-Depth Gains}
\label{app:recovery}

We measure how much of the BPB improvement from physical-depth
scaling is recovered by temporal thinking. Let $b_{\mathrm{base}}$
denote the BPB of the reference LRT without thought tokens,
$b_{\mathrm{deep}}$ that of a depth-scaled LRT, and
$b_{\mathrm{thought}}$ that of the corresponding thought-token
model. The recovery percentage is
\begin{equation}
 \mathrm{Recovery}
 =
 \frac{b_{\mathrm{base}}-b_{\mathrm{thought}}}
      {b_{\mathrm{base}}-b_{\mathrm{deep}}}
 \times 100\%.
\end{equation}

For doubling depth, the results in Table~\ref{tab:main} give
\begin{align}
 \mathrm{Recovery}_{16}
 &= \frac{0.798-0.783}{0.798-0.780}
 \times100\%
 = 83.3\%,\\
 \mathrm{Recovery}_{20}
 &= \frac{0.762-0.742}{0.762-0.741}
 \times100\%
 \approx 95.2\%.
\end{align}
For tripling depth, the corresponding recovery percentages
are 81.5\% and 83.9\%.

These quantities are specific to the training budget of
Table~\ref{tab:main}. Recovery at another budget requires the
reference, depth-scaled, and thought-token models to be
evaluated at that budget; the absolute gap between two models
alone does not determine the recovery percentage.

Parameter savings are calculated from the total-parameter
ratios in Table~\ref{tab:main}. One thought token uses
48.0\% and approximately 48.5\% fewer parameters than
double-depth LRT at the two reference sizes. Two thought
tokens use approximately 64.9\% and 65.3\% fewer parameters
than triple-depth LRT.

\section{Additional Experimental Details}
\label{app:details}

\begingroup
\color{black}
\subsection{Training FLOPs and Budget Matching}
\label{app:compute}

Table~\ref{tab:compute_breakdown} summarizes the computation of the five methods in the scaling comparison. Additional physical layers increase the backbone parameter count, whereas loops and refinement reuse parameters. Thought tokens also reuse parameters but lengthen the attention sequence. The columns describe these separate contributions; multiplying their entries is not an exact FLOP calculation.

\begin{table}[!t]
\centering
\color{black}
\small
\setlength{\tabcolsep}{6pt}
\renewcommand{\arraystretch}{1.15}
\caption{\textbf{Where training compute is spent.} Backbone parameters are normalized to the reference $L$-layer stack, excluding embeddings, the output head, and feedback projections. LRT feedback projections add approximately 5\% to the corresponding model's parameters and are shared across thought positions. Passes count complete backbone traversals per update, including loops or refinement as applicable; initialization is excluded. Expanded length is relative to 2,048 positions. Unique data is relative to LRT TT1 at the reported matched training budgets.}
\label{tab:compute_breakdown}
\begin{tabular}{@{}lcccc@{}}
\toprule
& Backbone & Mean passes & Expanded & Unique\\
Method & parameters & per update & length & data\\
\midrule
Vanilla Transformer, depth $2\times$ & $2\times$ & $1$ & $1\times$ & $\approx2\times$\\
LRT, depth $2\times$ & $2\times$ & $2$ & $1\times$ & $1\times$\\
Loop Transformer, 2 loops & $1\times$ & $2$ & $1\times$ & $\approx2\times$\\
PonderLM-2, 1 thought token & $1\times$ & $2$ & $2\times$ & $1\times$\\
LRT, 1 thought token & $1\times$ & $2$ & $2\times$ & $1\times$\\
\bottomrule
\end{tabular}
\end{table}

\paragraph{Pass sampling.}
All LRT and PonderLM-2 runs sample one $P\in\{1,2,3,4\}$ per batch with probabilities $(0.4,0.3,0.2,0.1)$, respectively. Thus $\mathbb{E}[P]=2$ exactly, although a finite run can have a slightly different realized mean. $P$ counts refinement passes and excludes initialization. Both $K$ and $P$ are shared across the sequences in a batch.

\paragraph{Matching the scaling budgets.}
For each model $m$, we calculate its average forward-and-backward backbone FLOPs per optimizer step, $\bar c_m$, using its depth, active computation, pass distribution, batch size, and expanded sequence length. Attention is included at the configured sequence length and window: expanding 2,048 positions to 4,096 doubles token-wise computation, but full-attention computation grows approximately fourfold. Thus the $2\times$ length entries in Table~\ref{tab:compute_breakdown} are not $2\times$ FLOP multipliers. The number of training steps is
\begin{equation}
 S_m\simeq\frac{C_{\mathrm{budget}}}{\bar c_m},
 \label{eq:training_budget}
\end{equation}
rounded to an integer, where $C_{\mathrm{budget}}$ is the corresponding vanilla double-depth reference budget. Initialization is outside this backbone-compute accounting.

The unique-data column records the half/twice relationship in the reported scaling runs. In Table~\ref{tab:scaling_full}, for example, vanilla double-depth and Loop2 at $D=20$ are compared with LRT TT1 at $D=10$, and their $D=160$ entries with LRT TT1 at $D=80$. Entries at the same $D$ instead have the same unique-data budget. Table~\ref{tab:main} reports results at $D=10$ and does not impose one common FLOP budget across all rows.

\paragraph{Fixed versus sampled $K$.}
Table~\ref{tab:variable_thinking} matches each sampled model to its counterpart trained at fixed $K=K_{\max}$ separately. Within a pair, both runs use the same expanded sequence length, batch size in sequences, $P$ distribution, optimizer-step count, and unique-data corpus. For fixed $K=1$ versus sampled $K\in\{0,1\}$, the executed length is always 4,096: $K=1$ uses 2,048 vocabulary and 2,048 thought positions, while $K=0$ uses 4,096 vocabulary positions. The sequence computation therefore remains the same under our backbone-FLOP accounting, so the same number of updates matches expected FLOPs. The $K_{\max}=2$ pair follows the same rule at 6,144 expanded positions.

Smaller sampled $K$ values process more vocabulary tokens per update. We supply these additional exposures by revisiting the same corpus, preserving equal unique data within each pair. Thus cumulative exposure can differ even when unique data, steps, and expected FLOPs are matched. Thought positions and repeated refinement of a batch do not count as additional unique data. No common training budget is imposed across all rows of Table~\ref{tab:variable_thinking}.
\endgroup

\subsection{Other Training and Evaluation Settings}

\paragraph{Inherited data and optimization settings.}
We use the recipe in Appendix~A of \citet{huang2026latent} (arXiv:2605.26797v2). Table~\ref{tab:training_recipe} records the inherited settings. The present study uses the MoE backbones and final-layer feedback specified in Section~\ref{sec:method}, with sampled full-sequence refinement rather than the cited paper's default interleaved schedule. Thought-token initialization has no direct loss; supervised refinement losses are averaged with gradients retained across passes.

\begin{table}[t]
\centering
\small
\setlength{\tabcolsep}{3pt}
\renewcommand{\arraystretch}{1.12}
\caption{Reference training recipe from \citet{huang2026latent}, Appendix~A. Sequence length and batch tokens below describe the vocabulary-only reference configuration; Appendix~\ref{app:compute} specifies expanded lengths and batching for fixed--variable comparisons.}
\label{tab:training_recipe}
\begin{tabularx}{\linewidth}{@{}>{\raggedright\arraybackslash}p{0.25\linewidth}>{\raggedright\arraybackslash}X@{}}
\toprule
Setting & Value\\
\midrule
Corpus & FineWeb-Edu~\citep{lozhkovfineweb} 100BT, nanochat pre-shuffled release\\
Validation & Final shard held out\\
Tokenizer & nanochat BPE; vocabulary 32,768\\
Reference sequence length & 2,048 vocabulary tokens\\
Reference global batch & $2^{19}$ vocabulary tokens/update\\
Training hardware & 8 H100 GPUs; BF16\\
Optimizer & MuonAdamW\\
Reference learning rates & Matrices: 0.02; embeddings/value embeddings: 0.3; head: 0.004; scalars: 0.5 (residual coefficients: additional $0.01\times$)\\
Adam betas & $(0.8,0.95)$; input-skip: $(0.96,0.95)$\\
LR scaling & Embedding/value-embedding/head rates: $(d/768)^{-1/2}$; all rates: $\sqrt{B_{\mathrm{vocab}}/2^{19}}$\\
LR schedule & No warmup; linear decay to zero over final half\\
Muon momentum & $0.85\rightarrow0.95$ over 300 steps\\
Weight decay & Depth factor $(12/L)^2$; annealed to zero\\
\bottomrule
\end{tabularx}
\end{table}

\paragraph{Evaluation.}
BPB divides summed target-token negative log-likelihood by $\ln 2$ times target UTF-8 bytes. Thought positions add no target bytes. Evaluation uses the sequential recurrence in Section~\ref{sec:setup}.

Sliding-window sizes are measured in expanded vocabulary-and-thought positions. At a fixed window size, increasing the thought count therefore reduces the number of vocabulary positions contained in that window.

\paragraph{Decoding latency.}
Following Appendix~D.5 of \citet{huang2026latent}, serving uses one H100-80GB, BF16, PyTorch 2.9, FlashAttention-3~\citep{shah2024flashattention}, batch 128, and 1,024 prompt plus 1,024 generated vocabulary tokens. For each model, we first normalize decoding latency to the vanilla baseline at the same reference size, then average the two normalized values from the 16L and 20L settings. This arithmetic mean is the single latency column in Table~\ref{tab:main}. Each output-token interval includes all thought steps ($K=1$ for TT1; $K=2$ for TT2); sampling $K$ during training does not reduce this inference cost.

\paragraph{Ablation controls.}
All ablations in Table~\ref{tab:ablation} use data ratio 10, normalized to the vanilla double-depth reference as defined in Appendix~\ref{app:results}. The full model uses training with variable thought steps with $K\in\{0,1\}$ and is evaluated at $K=1$.
The cross-token and KV-feedback ablations each remove one
feedback component while retaining training with variable thought steps.
The training ablation instead fixes $K=1$ while retaining
both feedback components.

\section{Variability Across Training Seeds}
\label{app:std}

Table~\ref{tab:main} reports mean validation BPB over three
training seeds. Table~\ref{tab:std} reports the corresponding
sample standard deviations at data ratio 10.
For three runs, the sample standard deviation is calculated as
\begin{equation}
 s_b=\sqrt{\frac{1}{n-1}\sum_{i=1}^{n}(b_i-\bar b)^2},
 \qquad n=3,
\end{equation}
where $b_i$ is the BPB of seed $i$ and $\bar b$ is its mean
across seeds. These standard deviations describe variation
across trained models, not uncertainty across validation tokens. Seed counts and variability for scaling and ablation runs are not inferred from the main table; they require separate reporting. In particular, the final 0.001 BPB scaling gap should not be interpreted as statistical equivalence.

\begin{table}[!t]
\centering
\small
\setlength{\tabcolsep}{3pt}
\renewcommand{\arraystretch}{1.12}
\caption{\textbf{Standard deviations of validation BPB at data ratio 10.}
Sample standard deviations across the three training seeds used for Table~\ref{tab:main}, reported separately for the 16L and 20L reference sizes.}
\label{tab:std}
\begin{tabularx}{\linewidth}{@{}>{\raggedright\arraybackslash}Xcc@{}}
\toprule
Model & 16L std & 20L std\\
\midrule
Transformer, depth $1\times$ & 0.0002 & 0.0002\\
Transformer, depth $2\times$ & 0.0005 & 0.0005\\
Transformer, depth $3\times$ & 0.0003 & 0.0001\\
\midrule
Loop Transformer, 2 loops & 0.0001 & 0.0006\\
Loop Transformer, 3 loops & 0.0006 & 0.0003\\
Loop Transformer, 4 loops & 0.0004 & 0.0002\\
\midrule
PonderLM-2, 1 thought token & 0.0003 & 0.0004\\
PonderLM-2, 2 thought tokens & 0.0002 & 0.0005\\
\midrule
LRT, depth $1\times$ & 0.0005 & 0.0001\\
LRT, depth $2\times$ & 0.0001 & 0.0003\\
LRT, depth $3\times$ & 0.0006 & 0.0006\\
\midrule
LRT TT1 & 0.0003 & 0.0002\\
LRT TT2 & 0.0004 & 0.0004\\
\bottomrule
\end{tabularx}
\end{table}

\end{document}